\documentclass[letterpaper, 10 pt, conference]{ieeeconf}  

\IEEEoverridecommandlockouts                              

\usepackage{graphics} 
\usepackage{epsfig} 
\usepackage{mathptmx} 
\usepackage{times} 
\usepackage{amsmath} 
\usepackage{amssymb}  
\usepackage{graphicx}
\usepackage{capt-of}
\usepackage{balance}
\usepackage{booktabs}
\usepackage{url}

\title{\LARGE \bf
MVG-WAM: Multiple View Geometry-Aware World-Action Modeling for Robotic Manipulation
}

\author{
    Wenbo Chen$^{1,*}$,
    Tianfu Li$^{1,*}$,
    Haoxuan Xu$^{2,*}$,
    Zhihao Cao$^{3}$,
    Zhenghan Chen$^{4}$,
    Zhengming Zhu$^{5}$,\\
    Zizhou Luo$^{6}$,
    Guosheng Yang$^{1}$,
    Yuan Liu$^{2}$,
    Lujia Wang$^{1}$,
    Wen Chen$^{7}$,
    Haoang Li$^{1}$%
    \thanks{$^{*}$Wenbo Chen, Tianfu Li, and Haoxuan Xu contributed equally to this work.}%
    \thanks{$^{1}$The Hong Kong University of Science and Technology (Guangzhou).}%
    \thanks{$^{2}$The Hong Kong University of Science and Technology.}%
    \thanks{$^{3}$ETH Zurich.}%
    \thanks{$^{4}$Zhejiang University.}%
    \thanks{$^{5}$EPFL.}%
    \thanks{$^{6}$University of Zurich.}%
    \thanks{$^{7}$The Chinese University of Hong Kong.}%
}

\IEEEaftertitletext{%
    \vspace{-0.5em}
    \begin{center}
        \includegraphics[width=\textwidth]{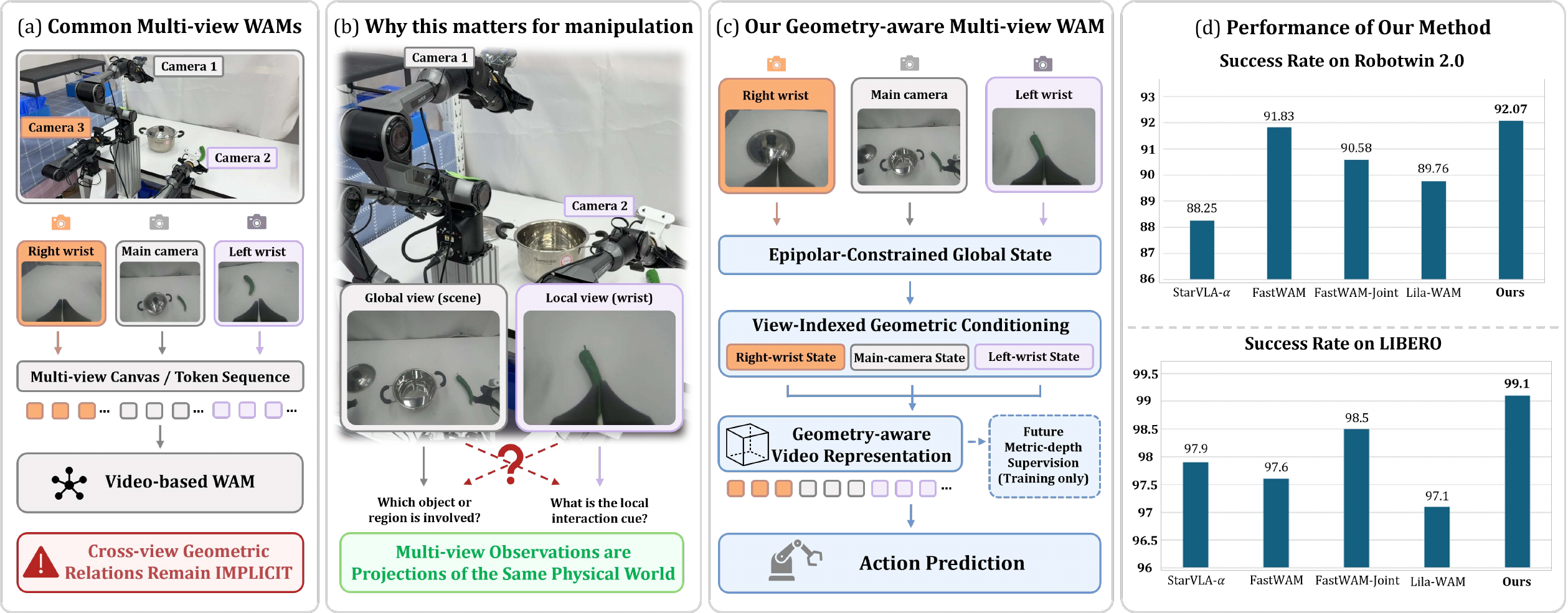}
        \captionof{figure}{
        \textbf{Motivation and conceptual overview of our MVG-WAM.}
        (a) Common multi-view interfaces tile images or concatenate tokens,
        leaving calibrated cross-view geometric relations implicit.
        (b) Scene and wrist views provide complementary global and local
        evidence about the same physical world.
        Manipulation therefore requires relating these observations to reason about
        object interaction.
        (c) MVG-WAM structures the representation of synchronized multi-view observations
        through an epipolar-constrained global state and view-indexed geometric conditioning,
        producing a geometry-aware video representation for action prediction.
        Future metric-depth supervision metrically grounds the geometry-aware
        video representation during training and is removed at deployment.
        (d) Among methods without additional embodied pretraining,
        MVG-WAM achieves strong average success rates
        on both LIBERO and RoboTwin~2.0.}
        \label{fig:teaser}
    \end{center}
    \vspace{-0.5em}
}

\begin{document}

\maketitle
\thispagestyle{empty}
\pagestyle{empty}

\begin{abstract}
World--Action Models (WAMs) couple visual dynamics with action prediction, bringing the rich priors of pretrained video models to robotic manipulation.
However, their multi-view interfaces typically tile images or concatenate tokens, leaving the geometric relationships among synchronized cameras implicit.
This makes it harder to connect global scene context with the local geometry required for interaction.
We introduce the Multi-View Geometry-Aware World--Action Model (MVG-WAM), which organizes these observations as related projections of one physical world rather than separate images on a canvas.
Our model combines an epipolar-constrained global state with view-indexed geometric states jointly inferred from synchronized observations.
Camera-aware routing supplies each video region with its corresponding geometric context and the shared global state, explicitly structuring the representation used for action prediction.
We further ground the geometry-aware representation in metric scale through multi-horizon future-depth supervision, without requiring depth decoding during action rollout.
MVG-WAM achieves average success rates of 99.1\% on LIBERO and 92.07\% on RoboTwin~2.0, demonstrating competitive performance across both benchmarks.
Real-world experiments on Cobot Magic further demonstrate a 91.3\% success rate across 150 trials spanning three manipulation tasks. Our project page are available at \url{https://bobc-123.github.io/MVG-WAM/}.
\end{abstract}

\section{INTRODUCTION}

World--Action Models (WAMs) jointly model future visual dynamics and robot actions, allowing control to benefit from predictive representations of the environment~\cite{li2025uva,zhu2025uwm}.
This paradigm has attracted growing attention for robotic manipulation.
To exploit large-scale spatiotemporal priors, many recent WAMs build on pretrained video-generation backbones~\cite{bi2025motus,ye2026dreamzero,ma2026dit4dit}.
While this inheritance provides strong visual dynamics priors, it also carries over the representation interface of video generation.

This interface becomes problematic in multi-view manipulation.
Robots commonly observe the workspace through a scene camera and multiple wrist cameras.
The scene view provides global workspace context, while wrist views reveal local geometry around the gripper and contact regions.
Crucially, these observations are synchronized projections of the same physical world.
Existing WAMs often accommodate them by tiling multiple views on a single video canvas or concatenating their visual tokens~\cite{ye2026dreamzero,yuan2026fastwam}.
As illustrated in Fig.~\ref{fig:teaser}(a), these interfaces preserve each camera's visual content but leave calibrated cross-view relations implicit.
Manipulation nevertheless requires linking global scene context to local interaction geometry (Fig.~\ref{fig:teaser}(b)).
The model must therefore recover cross-view correspondences from the training signal while simultaneously learning visual dynamics and actions.
This additional burden can weaken geometric coherence across views and, in turn, spatial reasoning for action prediction.

To this end, we introduce the \textbf{Multi-View Geometry-Aware World--Action Model (MVG-WAM)}.
Our design follows a simple principle: \emph{multiple camera observations should be represented as geometrically related projections of one physical world, rather than merely as multiple images on a video canvas}.
We therefore explicitly organize the multi-view representation around the physical relations among synchronized camera observations.

As outlined in Fig.~\ref{fig:teaser}(c), MVG-WAM realizes this principle through three complementary mechanisms.
First, we construct a global state under explicit epipolar constraints, allowing information to be aggregated across geometrically related views.
Second, a geometry foundation model jointly encodes the synchronized observations into correlated per-view states.
We group the video tokens according to their camera regions.
Each group is conditioned on the corresponding per-view state.
This preserves view alignment while injecting multi-view geometric context.
Together, the first two mechanisms provide relational multi-view geometry
through calibrated cross-view aggregation and view-indexed conditioning.
We complement this structure with multi-horizon metric-depth supervision,
which anchors the geometry-aware video representation to physical scale.
This metric grounding encourages the latent dynamics to preserve physically
meaningful geometry as the scene evolves under actions.

Our contributions are fourfold:
\begin{itemize}
    \item We formulate multi-view observations in WAMs as geometrically related projections of a shared physical world, and introduce an epipolar-constrained global state to model their shared structure.

    \item We introduce view-indexed geometric conditioning from jointly encoded multi-view features, linking each camera region of the video representation to its corresponding geometric state.

    \item We metrically ground the geometry-aware video representation through multi-horizon future-depth supervision, encouraging physically scaled geometry across views and time without requiring depth decoding at deployment.

    \item Without additional embodied pretraining, our MVG-WAM achieves competitive performance on both LIBERO and RoboTwin~2.0, together with strong real-world manipulation results.
\end{itemize}

\section{Related Work}

\subsection{World--Action Modeling}

World--Action Models (WAMs) jointly model visual dynamics and robot actions.
UVA~\cite{li2025uva} and UWM~\cite{zhu2025uwm} established unified video--action learning through shared representations or diffusion formulations.
Subsequent methods, including Motus~\cite{bi2025motus}, mimic-video~\cite{pai2025mimicvideo}, LingBot-VA~\cite{li2026lingbotva}, and DreamZero~\cite{ye2026dreamzero}, further exploit large-scale video priors for robot control.
FastWAM~\cite{yuan2026fastwam} shows that video co-training remains useful even without future-frame generation at deployment.
Together, these works highlight video representations as an effective basis for action learning, while multi-view observations are still commonly handled through generic video interfaces that leave calibrated cross-view relations implicit.

\subsection{Geometry-Aware Robot Learning}

Geometry has increasingly been introduced into robot learning to improve spatial reasoning and manipulation accuracy.
For VLA policies, GeoAware-VLA~\cite{abouzeid2025geoawarevla} incorporates geometric visual features, while Spatial Forcing~\cite{li2025spatialforcing} and ROCKET~\cite{sun2026rocket} transfer spatial priors from pretrained 3D models.
A similar trend appears in WAMs~\cite{he2026sgwamsemantic}:
GeoSem-WAM~\cite{ma2026geosemwam} incorporates geometric and semantic cues, X-WAM~\cite{guo2026xwam} predicts multi-view RGB-D futures, and WAM4D~\cite{li2026wam4d} uses future depth as geometric supervision.
MECo-WAM~\cite{zhang2026mecowam} and SG-WAM~\cite{zhao2026sgwam} further explore geometric priors for action-conditioned representations.
These works mainly study \emph{what geometric information} should be encoded or predicted; our focus is instead on \emph{how synchronized camera observations should be geometrically related within a multi-view WAM}.

\subsection{Multi-View Geometric Modeling}

Multi-view geometry provides explicit mechanisms for relating different projections of a shared 3D scene.
MVDiff~\cite{bourigault2024mvdiff}, EpiDiff~\cite{huang2024epidiff}, and MVISTA-4D~\cite{wang2026mvista} use epipolar constraints or geometry-aware feature interaction to improve cross-view correspondence.
In parallel, geometry foundation models such as VGGT~\cite{wang2025vggt}, VGGT-$\Omega$~\cite{wang2026vggtomega}, and Depth Anything~3~\cite{lin2025da3} provide strong geometric priors from multi-view observations.
These directions are complementary: calibrated geometry constrains where cross-view information is exchanged, while learned geometric representations capture scene structure across views.
MVG-WAM brings these ideas into world--action modeling, organizing synchronized observations as geometrically related projections rather than independent regions of a video interface.

\begin{figure*}[t]
    \centering
    \includegraphics[width=\textwidth]{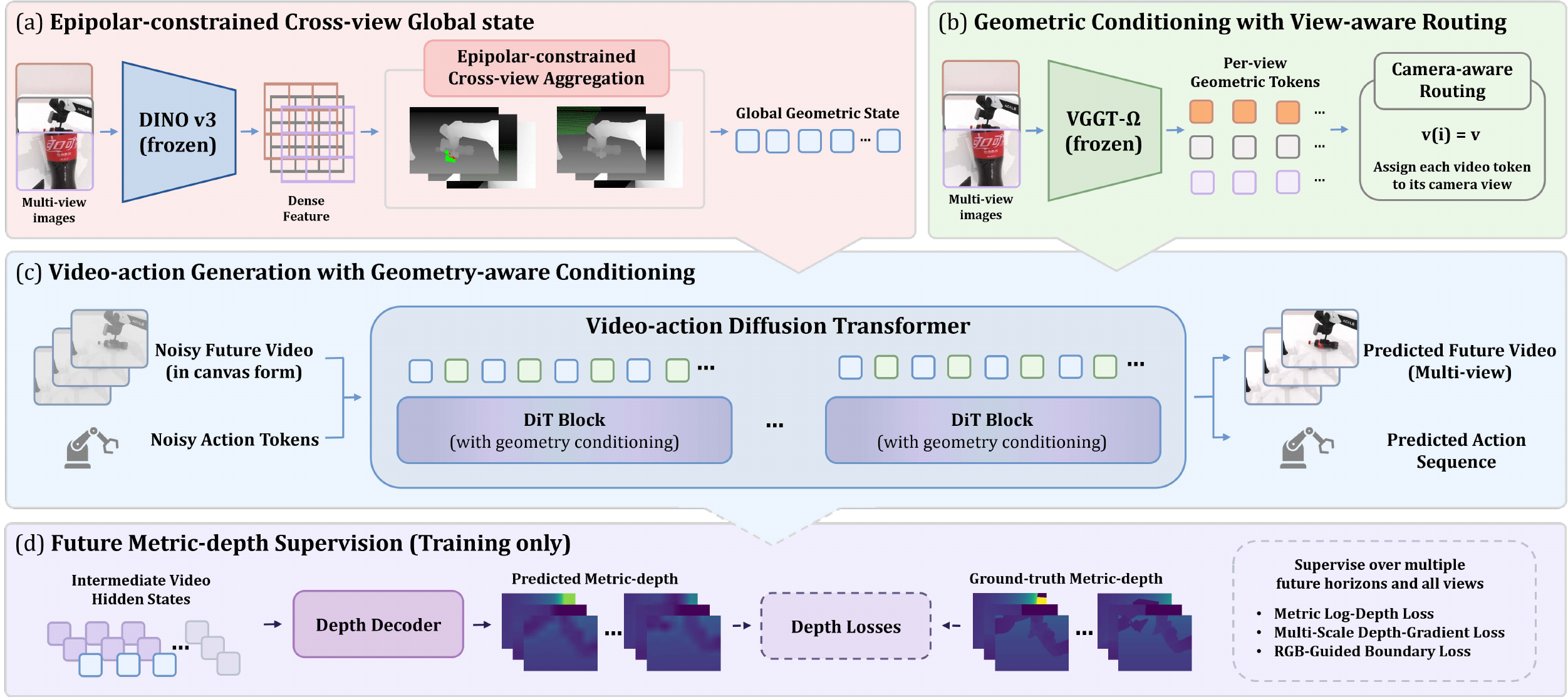}
    \caption{
    \textbf{Architecture of our MVG-WAM.}
    (a) Dense DINOv3 features are aggregated under calibrated epipolar
    constraints to provide cross-view geometric context.
    (b) Compact VGGT-$\Omega$ registers form view-indexed geometric states,
    which are routed to the corresponding camera regions.
    (c) The global and view-indexed states condition the video stream of the
    video--action diffusion transformer, while action tokens access the
    geometry-enhanced video representation through mixed attention.
    (d) During training, future metric-depth supervision provides metric-scale
    grounding for intermediate denoising features across views and horizons;
    the depth decoder is not used during action rollout.
    }
    \label{fig:method_overview}
\end{figure*}

\section{Methodology}
\label{sec:method}

\subsection{WAM Formulation and Method Overview}
\label{sec:method_overview}

We consider a video-based World--Action Model with a video expert and an action expert.
At time $t$, the robot receives a language instruction $c$, a proprioceptive state $s_t$, and synchronized observations from $V$ calibrated cameras:
\begin{equation}
    \mathcal{O}_t
    =
    \left\{
        I_t^v,K_t^v,T_t^v
    \right\}_{v=1}^{V}.
    \label{eq:observation}
\end{equation}
Here, $I_t^v$ is the RGB observation and $K_t^v$ is the intrinsic matrix.
$T_t^v\in SE(3)$ maps world-frame points to camera $v$.
The synchronized images are arranged into a multi-view canvas, denoted by $X_t$.

During training, the model predicts a future canvas sequence
$X_{t+1:t+H_v}$ and an action trajectory
$\mathbf a_{t:t+H_a-1}$.
A video VAE maps the canvas sequence into latent video tokens.
The current canvas remains a clean first-frame condition.
Future video latents and action tokens are independently perturbed at diffusion timesteps $\tau_v$ and $\tau_a$.
The two experts then predict their corresponding flow-matching targets.

Action tokens attend to the full video sequence through mixed attention.
Language and proprioception provide additional conditioning.
The base training objective contains a video loss
$\mathcal L_{\mathrm{vid}}$ and an action loss
$\mathcal L_{\mathrm{act}}$.

Figure~\ref{fig:method_overview} summarizes the architecture of MVG-WAM.
As shown in Fig.~\ref{fig:method_overview}(a), calibrated epipolar geometry
constrains cross-view feature aggregation into a compact global state.
Figure~\ref{fig:method_overview}(b) introduces view-indexed geometric states
and camera-aware routing for the corresponding canvas regions.
These two geometric conditions are integrated into the video--action backbone
in Fig.~\ref{fig:method_overview}(c).
Finally, Fig.~\ref{fig:method_overview}(d) provides metric-scale grounding
through future-depth supervision during training.

MVG-WAM retains the video interface while injecting calibrated
cross-view and view-indexed geometric structure into its latent representation.

\subsection{Epipolar-Constrained Cross-View Global State}
\label{sec:epipolar_global}

The video backbone operates on compact VAE latents optimized for efficient generation.
Such strongly compressed representations can limit information capacity and fine-grained visual structure~\cite{zheng2026rae}.
We therefore perform geometry-constrained cross-view retrieval in a dense feature space.

For each current observation, we extract a dense DINOv3 patch feature map~\cite{simeoni2025dinov3}.
We denote the projected feature map by
$S_t^v\in\mathbb{R}^{H_f\times W_f\times d_s}$.
All cameras share the same feature projector.
A learned view embedding identifies the source camera of each feature map.

The camera calibration must remain consistent with image preprocessing.
Let $A_v$ map coordinates from the original image to the dense encoder input.
For readability, we omit the time index from camera quantities within this subsection.
The feature-space geometry is
\begin{equation}
\begin{aligned}
    \bar K_v
    &=
    A_vK_v,\\
    T_{sq}
    &=
    T_sT_q^{-1}
    =
    [R_{sq}\mid t_{sq}],\\
    E_{sq}
    &=
    [t_{sq}]_{\times}R_{sq},\\
    F_{qs}
    &=
    \bar K_s^{-\top}
    E_{sq}
    \bar K_q^{-1}.
\end{aligned}
\label{eq:epipolar_geometry}
\end{equation}
For a query patch center $\tilde p$, the corresponding source observation must lie on
$\ell_{qs}(p)=F_{qs}\tilde p$.
The epipolar line restricts correspondence search to a geometrically admissible region, within which attention selects candidate features.

We clip each epipolar line to the valid source feature region.
A narrow band is sampled along the resulting in-image segment.
The continuous locations are converted to feature-grid coordinates, and source features are obtained by bilinear interpolation.

A query is marked valid, $m_{qs}(p)=1$, when the epipolar line
intersects the source feature region and contains at least one valid sampled candidate.

Following epipolar-constrained multi-view interaction~\cite{huang2024epidiff}, the query feature attends to its sampled candidates using masked multi-head attention.
The attention includes a learned candidate-position bias.
A learnable null candidate provides a no-match option.
The resulting attended feature is denoted by $U_{qs}(p)$.

We compute a confidence value
$c_{qs}(p)=
\sigma(g([S_t^q(p);U_{qs}(p)]))$.
The directed cross-view residual is
\begin{equation}
    \Delta S_{qs}(p)
    =
    m_{qs}(p)\,
    c_{qs}(p)\,
    U_{qs}(p).
    \label{eq:epipolar_update}
\end{equation}
Queries without valid real candidates receive an exact zero update.
The null candidate and confidence gate reduce the effect of unsupported matches.

We apply the same operator to all valid ordered camera pairs.
Each view can therefore act as both query and source.
Let
$\mathcal V_q(p)=
\{s\neq q\mid m_{qs}(p)=1\}$
be the valid source-view set.
The enhanced query feature is
\begin{equation}
    \widetilde S_t^q(p)
    =
    S_t^q(p)
    +
    \frac{1}{|\mathcal V_q(p)|}
    \sum_{s\in\mathcal V_q(p)}
    \Delta S_{qs}(p).
    \label{eq:epipolar_aggregation}
\end{equation}
The residual term is set to zero when $\mathcal V_q(p)$ is empty.

We concatenate the enhanced features into
$M_t^{\mathrm{epi}}
=
[\widetilde S_t^1;\ldots;\widetilde S_t^V]$.
A learnable query bank pools this memory into a compact residual:
\begin{equation}
\begin{aligned}
    \Delta G_t^{\mathrm{epi}}
    &=
    \operatorname{CAStack}
    \left(
        Q_{\mathrm{epi}},
        M_t^{\mathrm{epi}}
    \right),\\
    G_t
    &=
    G_t^{\mathrm{base}}
    +
    \alpha_{\mathrm{epi}}
    \Delta G_t^{\mathrm{epi}}.
\end{aligned}
\label{eq:epipolar_global}
\end{equation}
$\operatorname{CAStack}$ denotes a small stack of multi-head cross-attention and feed-forward blocks.
$G_t^{\mathrm{base}}$ is the compact geometry memory introduced in Sec.~\ref{sec:view_condition}.
The residual gate $\alpha_{\mathrm{epi}}$ is learned during training.

Query pooling compresses the epipolar-enhanced dense features
into a compact cross-view memory $\Delta G_t^{\mathrm{epi}}$.

\subsection{View-Indexed Geometric Conditioning}
\label{sec:view_condition}

The epipolar branch provides dense cross-view evidence.
We additionally require compact camera-level context for each input view.
To this end, a frozen geometry foundation model jointly processes the synchronized observations.
We instantiate this encoder with VGGT-$\Omega$~\cite{wang2026vggtomega}.

For every input view, VGGT-$\Omega$ produces a camera-and-register token sequence.
We discard the camera token and retain the compact view-indexed register bank $R_t^v$.

A shared adapter maps the registers into the WAM feature space.
A learnable query bank then summarizes all adapted states:
\begin{equation}
\begin{aligned}
    P_t^v
    &=
    \mathcal A_{\mathrm{reg}}
    (R_t^v)
    +
    e_v,\\
    P_t
    &=
    [P_t^1;\ldots;P_t^V],\\
    G_t^{\mathrm{base}}
    &=
    \operatorname{CAStack}
    (Q_{\mathrm{geo}},P_t).
\end{aligned}
\label{eq:geometry_states}
\end{equation}
The embedding $e_v$ identifies the camera stream,
while current camera geometry enters through the calibrated epipolar branch.

$Q_{\mathrm{geo}}$ and $Q_{\mathrm{epi}}$ use matched slot counts and dimensions,
allowing the epipolar branch to act as a residual correction.


We next associate latent video tokens with their camera regions.
Let $\nu(i)$ denote the region containing the latent-grid center of token $i$.
This spatial association is repeated across all latent temporal groups.
The current state $P_t^v$ therefore conditions all temporal tokens associated with region $v$.

The routing assigns each latent token to the camera region containing its latent-grid center.

Let $\bar h_i^{(l)}$ denote a video token after mixed attention at layer $l$.
We define the view and global conditions by
\begin{equation}
\begin{aligned}
    C_{i,v}^{(l)}
    &=
    \operatorname{CA}_{v}^{(l)}
    \left(
        \bar h_i^{(l)},
        P_t^{\nu(i)}
    \right),\\
    C_{i,g}^{(l)}
    &=
    \operatorname{CA}_{g}^{(l)}
    \left(
        \bar h_i^{(l)},
        G_t
    \right).
\end{aligned}
\end{equation}
The conditioned token is
\begin{equation}
    \widehat h_i^{(l)}
    =
    \bar h_i^{(l)}
    +
    \alpha_v^{(l)}
    C_{i,v}^{(l)}
    +
    \alpha_g^{(l)}
    C_{i,g}^{(l)}.
    \label{eq:view_conditioning}
\end{equation}
The two branches use independent cross-attention parameters.
Their layer-specific residual gates are initialized to zero.

Geometric conditioning is inserted after mixed attention and before
text cross-attention and the block FFN throughout the video expert.
Action tokens subsequently access the geometry-enhanced video representation
through mixed-attention layers.

\begin{table*}[t]
    \centering
    \caption{
    Success rate (\%) on LIBERO and RoboTwin~2.0.
    Emb. PT. indicates additional embodied/robot-data pretraining
    before target-benchmark training.
    Bold and underlined values denote the highest and second-highest values
    in each benchmark column, respectively.
    }
    \label{tab:sim_main}
    \begingroup
    \footnotesize
    \setlength{\tabcolsep}{2.7pt}
    \renewcommand{\arraystretch}{1.06}
    \begin{tabular*}{\textwidth}{@{\extracolsep{\fill}}lccccccccc@{}}
        \toprule
        & & \multicolumn{5}{c}{LIBERO}
        & \multicolumn{3}{c}{RoboTwin 2.0} \\
        \cmidrule(lr){3-7}\cmidrule(l){8-10}
        Method & Emb. PT.
        & Spatial & Object & Goal & Long & Avg.
        & Clean & Rand. & Avg. \\
        \midrule

        $\pi_{0.5}$~\cite{black2025pi05,yuan2026fastwam}
        & Yes
        & 98.8 & 98.2 & 98.0 & 92.4 & 96.9
        & 82.74 & 76.76 & 79.75 \\

        X-VLA~\cite{zheng2025xvla,bi2025motus}
        & Yes
        & 98.2 & 98.6 & 97.8 & 97.6 & 98.1
        & 72.80 & 72.84 & 72.82 \\

        ABot-M0~\cite{yang2026abotm0}
        & Yes
        & 98.8 & \underline{99.8} & \textbf{99.0} & 96.6 & \underline{98.6}
        & 80.42 & 81.16 & 80.79 \\

        StarVLA-$\alpha$~\cite{ye2026starvlaalpha}
        & No
        & 99.0 & \underline{99.8} & \underline{98.5} & 94.1 & 97.9
        & 88.20 & 88.30 & 88.25 \\

        \midrule

        Motus~\cite{bi2025motus,yuan2026fastwam}
        & Yes
        & 96.8 & \underline{99.8} & 96.6 & 97.6 & 97.7
        & 88.66 & 87.02 & 87.84 \\

        LingBot-VA~\cite{li2026lingbotva}
        & Yes
        & 98.5 & 99.6 & 97.2 & \textbf{98.5} & 98.5
        & \textbf{92.93} & 91.55 & \textbf{92.24} \\

        FastWAM~\cite{yuan2026fastwam}
        & No
        & 98.2 & \textbf{100.0} & 97.0 & 95.2 & 97.6
        & 91.88 & \textbf{91.78} & 91.83 \\

        FastWAM-Joint~\cite{yuan2026fastwam}
        & No
        & \underline{99.6} & 99.4 & 98.2 & 96.8 & 98.5
        & 90.84 & 90.32 & 90.58 \\

        LiLa-WAM~\cite{yang2026lilawam}
        & No
        & 98.0 & 98.8 & 97.2 & 94.2 & 97.1
        & 90.48 & 89.04 & 89.76 \\

        \midrule

        \textbf{MVG-WAM (Ours)}
        & \textbf{No}
        & \textbf{99.8} & \underline{99.8} & \textbf{99.0}
        & \underline{97.8} & \textbf{99.1}
        & \underline{92.54} & \underline{91.60} & \underline{92.07} \\

        \bottomrule
    \end{tabular*}
    \endgroup
\end{table*}

\subsection{Future Metric-Depth Supervision}
\label{sec:depth_supervision}

Epipolar aggregation and view-indexed conditioning establish relational
geometry across synchronized observations.
To anchor this geometry-aware representation to physical scale, we supervise
intermediate denoising features with metric depth at multiple future horizons.
This metric grounding complements the cross-view structure by encouraging
the evolving video representation to preserve absolute spatial scale.

Let
$\Delta=\{\delta_1,\ldots,\delta_{H_d}\}$
denote the supervised future horizons.
For camera $v$ and horizon $\delta_h$, the target
$D_{t+\delta_h}^{v,*}$
is camera-axis depth in the corresponding future camera frame.
All targets are expressed in the same physical unit, providing a shared
metric reference for the view-indexed geometry across future horizons.

Metric depth is used as auxiliary training supervision,
while rollout relies only on RGB, proprioception, language, and calibration.

We collect complete post-block video states from a set of layers $\mathcal L$.
These states have passed through mixed attention, MVG conditioning, text cross-attention, and the block FFN.
The latent canvas is divided using the same camera-region layout as the conditioning module.

Each selected feature is normalized and projected to a common decoder dimension.
Multi-level features from the same camera region are fused by a shared convolutional decoder.
View and temporal-group embeddings distinguish the outputs.
The video diffusion timestep is injected through FiLM modulation~\cite{perez2018film}.

Let
$H_{\mathcal L}^v=
\{H_{\tau_v}^{(l),v}\}_{l\in\mathcal L}$
collect the selected features for view $v$.
The decoder produces log-depth and metric depth as
\begin{equation}
\begin{aligned}
    \widehat{\mathbf d}^{\,v}
    &=
    \mathcal D
    \left(
        H_{\mathcal L}^v,
        \tau_v,
        v
    \right),\\
    \widehat{\mathbf D}^{\,v}
    &=
    \exp
    \left(
        \operatorname{clip}
        (\widehat{\mathbf d}^{\,v})
    \right).
\end{aligned}
\label{eq:future_depth}
\end{equation}
The entries of $\widehat{\mathbf D}^{\,v}$ follow the horizon order in $\Delta$.
The clipping interval
$[d_{\min},d_{\max}]$
is defined in log-depth space.
Invalid pixels and unavailable horizons are excluded by validity masks.

Because the decoder reads denoising features produced from noised future latents,
the depth objective acts as metric grounding for the evolving representation
rather than as a standalone forecasting module.

The auxiliary loss contains three components.
$\mathcal L_{\log}$ is a Huber loss between predicted and target log-depth over valid pixels.
$\mathcal L_{\nabla}$ matches horizontal and vertical log-depth gradients at multiple spatial scales.
$\mathcal L_{\mathrm{bnd}}$ applies the gradient residual with an RGB-derived spatial weight.

The boundary weight is computed from the normalized grayscale gradient of the corresponding future ground-truth RGB frame.
For video timestep $\tau_v$, we use
$w(\tau_v)=
\max((1-\tau_v/T)^\gamma,w_{\min})$
to reduce supervision at high noise levels.

The depth and overall objectives are
\begin{equation}
\begin{aligned}
    \mathcal L_{\mathrm{aux}}
    &=
    \lambda_{\log}\mathcal L_{\log}
    +
    \lambda_{\nabla}\mathcal L_{\nabla}
    +
    \lambda_{\mathrm{bnd}}\mathcal L_{\mathrm{bnd}},\\
    \mathcal L_{\mathrm{depth}}
    &=
    \mathbb E
    \left[
        w(\tau_v)
        \mathcal L_{\mathrm{aux}}
    \right],\\
    \mathcal L
    &=
    \lambda_{\mathrm{vid}}\mathcal L_{\mathrm{vid}}
    +
    \lambda_{\mathrm{act}}\mathcal L_{\mathrm{act}}
    +
    \lambda_{\mathrm{dep}}\mathcal L_{\mathrm{depth}}.
\end{aligned}
\label{eq:training_objective}
\end{equation}

The geometry encoder, video VAE, and text encoder remain frozen.
The video and action experts are jointly optimized with the new MVG modules.
We use separate learning rates for pretrained and newly initialized parameters.

During rollout, the geometry representation is computed once from the
current synchronized observation and reused across denoising steps.
The auxiliary depth decoder is disabled at inference.

\section{Experiments}
\label{sec:experiments}

Our experiments address four questions:

\textbf{Q1}, how well does MVG-WAM perform in simulation and on a real robot?

\textbf{Q2}, how robust is it to distribution shifts?

\textbf{Q3}, which geometric components contribute to control?

\textbf{Q4}, how well does the learned representation support short-horizon prediction and metric multi-view geometry?

\subsection{Experimental Setup}
\label{sec:exp_setup}

\textbf{Data and evaluation.}
LIBERO~\cite{liu2023libero} contains 40 tasks across Spatial, Object, Goal, and Long, with 50 demonstrations per task.
RoboTwin~2.0~\cite{chen2025robotwin} uses 50 tasks with 50 clean and 500 randomized demonstrations per task.
We additionally record synchronized simulator depth for auxiliary supervision.
One policy is trained jointly across all LIBERO suites, and a separate policy covers all RoboTwin~2.0 tasks.
Following FastWAM~\cite{yuan2026fastwam}, we evaluate 50 trials per LIBERO task and 100 trials per RoboTwin~2.0 task and domain.

\textbf{Training and inference.}
The video backbone is initialized from Wan2.2~\cite{wang2025wan}.
The video and action experts are optimized end-to-end with the MVG modules, while the geometry encoder, video VAE, and text encoder remain frozen.
Training uses eight H100 GPUs, bfloat16, a global batch size of 128, and AdamW~\cite{loshchilov2019adamw}.
Learning rates are $5\times10^{-6}$ for pretrained parameters and $10^{-4}$ for newly introduced modules.
LIBERO and RoboTwin~2.0 are trained for 30k and 100k updates, respectively.
Inference uses ten denoising steps and 32-action chunks; the auxiliary depth decoder is not used during control.

\subsection{Simulation Benchmark Results}
\label{sec:exp_simulation}

To address Q1 in simulation, Table~\ref{tab:sim_main} compares MVG-WAM
with representative VLA and WAM policies, while distinguishing whether
additional embodied pretraining is used.

Without additional embodied pretraining, MVG-WAM achieves 99.1\% average
SR on LIBERO and 92.07\% on RoboTwin~2.0.
Compared with FastWAM-Joint, this corresponds to gains of 0.6 and
1.49 percentage points, respectively.

On RoboTwin~2.0, MVG-WAM also outperforms FastWAM by 0.24 points and remains
only 0.17 points below the embodied-pretrained LingBot-VA.
For reference, LingBot-VA reports 5.23 s under its RoboTwin~2.0 configuration~\cite{zhang2026wamrobustness},
while our implementation takes 0.74 s per 32-step chunk on an H100.
These results show that MVG-WAM attains competitive manipulation performance
without additional embodied pretraining, while maintaining substantially lower
inference latency than LingBot-VA.

The Clean--Randomized gap is 0.94 points.
Since randomized demonstrations are included during training,
this comparison does not constitute an OOD evaluation.

\subsection{Out-of-Distribution Generalization}
\label{sec:exp_ood}

To address Q2, we evaluate two complementary distribution shifts:
the unchanged LIBERO policy on LIBERO-Plus~\cite{fei2025liberoplus}, and clean-only RoboTwin~2.0 training followed by Clean2Random evaluation.

\textbf{LIBERO-Plus.}
Table~\ref{tab:libero_plus} compares our evaluation with published
OpenVLA and OpenVLA-OFT results~\cite{kim2024openvla,kim2025openvlaoft}.
MVG-WAM improves the pooled SR from 68.7\% for FastWAM-Joint to 70.4\%.
The largest gains over FastWAM-Joint occur under Camera and Background
shifts (+12.0 and +4.2 points), while performance decreases under
Language and Light perturbations.
The gains therefore do not extend uniformly across all shifts.

\begin{table}[t]
    \centering
    \caption{
    LIBERO-Plus success rate (\%) under seven distribution shifts.
    OpenVLA and OpenVLA-OFT results follow LIBERO-Plus~\cite{fei2025liberoplus};
    FastWAM-Joint~\cite{yuan2026fastwam} and MVG-WAM use our evaluation.
    Bold and underlined values denote the highest and second-highest
    values in each row, respectively.
    }
    \label{tab:libero_plus}
    \begingroup
    \footnotesize
    \setlength{\tabcolsep}{3pt}
    \renewcommand{\arraystretch}{1.06}
    \begin{tabular*}{\columnwidth}{@{\extracolsep{\fill}}lcccc@{}}
        \toprule
        Shift & OpenVLA & OpenVLA-OFT & FastWAM-Joint & \textbf{MVG-WAM} \\
        \midrule

        Camera
        & 0.8
        & \textbf{56.4}
        & 39.9
        & \underline{51.9} \\

        Robot
        & 3.5
        & 31.9
        & \underline{65.1}
        & \textbf{65.3} \\

        Language
        & 23.0
        & 79.5
        & \textbf{94.7}
        & \underline{90.1} \\

        Light
        & 8.1
        & \underline{88.7}
        & \textbf{92.1}
        & 88.4 \\

        Background
        & 34.8
        & \textbf{93.3}
        & 58.1
        & \underline{62.3} \\

        Noise
        & 15.2
        & \textbf{75.8}
        & 56.2
        & \underline{57.6} \\

        Layout
        & 28.5
        & 74.2
        & \underline{79.3}
        & \textbf{80.8} \\

        \midrule

        Overall
        & 15.6
        & \underline{69.6}
        & 68.7
        & \textbf{70.4} \\

        \bottomrule
    \end{tabular*}
    \endgroup
\end{table}

\textbf{RoboTwin~2.0 Clean2Random.}
Following the official Clean2Random setting, we train FastWAM-Joint and
MVG-WAM as multi-task policies on the same 2,500 clean demonstrations
(50 per task) without exposure to randomized RoboTwin~2.0 data.
Both policies are trained for 100k updates and evaluated over 100 trials
per task in each test domain.

Table~\ref{tab:clean2random} additionally includes the X-WAM, FastWAM,
and AHA-WAM results reported in the RoboTwin~2.0
leaderboard snapshot~\cite{community2026xpolicylab}.
MVG-WAM reaches 72.6\% C2C and 29.2\% C2R SR, improving over
FastWAM-Joint by 2.8 and 27.9 percentage points, respectively.
The remaining gap between C2C and C2R indicates that unseen randomized
conditions remain substantially more challenging.

\begin{table}[t]
    \centering
    \caption{
    RoboTwin~2.0 Clean2Clean (C2C), Clean2Random (C2R), and average
    success rates (\%).
    X-WAM, FastWAM, and AHA-WAM results follow the RoboTwin~2.0 leaderboard snapshot~\cite{community2026xpolicylab};
    FastWAM-Joint and MVG-WAM are evaluated in our clean-only setting.
    Bold and underlined values denote the highest and second-highest
    values in each row, respectively.
    }
    \label{tab:clean2random}
    \begingroup
    \footnotesize
    \setlength{\tabcolsep}{2.5pt}
    \renewcommand{\arraystretch}{1.08}

    \begin{tabular*}{\columnwidth}{
        @{\extracolsep{\fill}}lccccc@{}
    }
        \toprule
        & X-WAM & FastWAM & AHA-WAM & FastWAM-Joint & \textbf{MVG-WAM} \\
        \midrule

        C2C
        & 70.0
        & \textbf{77.8}
        & 64.3
        & 69.8
        & \underline{72.6} \\

        C2R
        & \underline{25.8}
        & 1.9
        & 3.2
        & 1.3
        & \textbf{29.2} \\

        Avg.
        & \underline{47.9}
        & 39.9
        & 33.8
        & 35.6
        & \textbf{50.9} \\

        \bottomrule
    \end{tabular*}

    \endgroup
\end{table}

\begin{figure*}[t]
    \centering
    \IfFileExists{figures/3.2.pdf}{%
        \includegraphics[width=\textwidth]{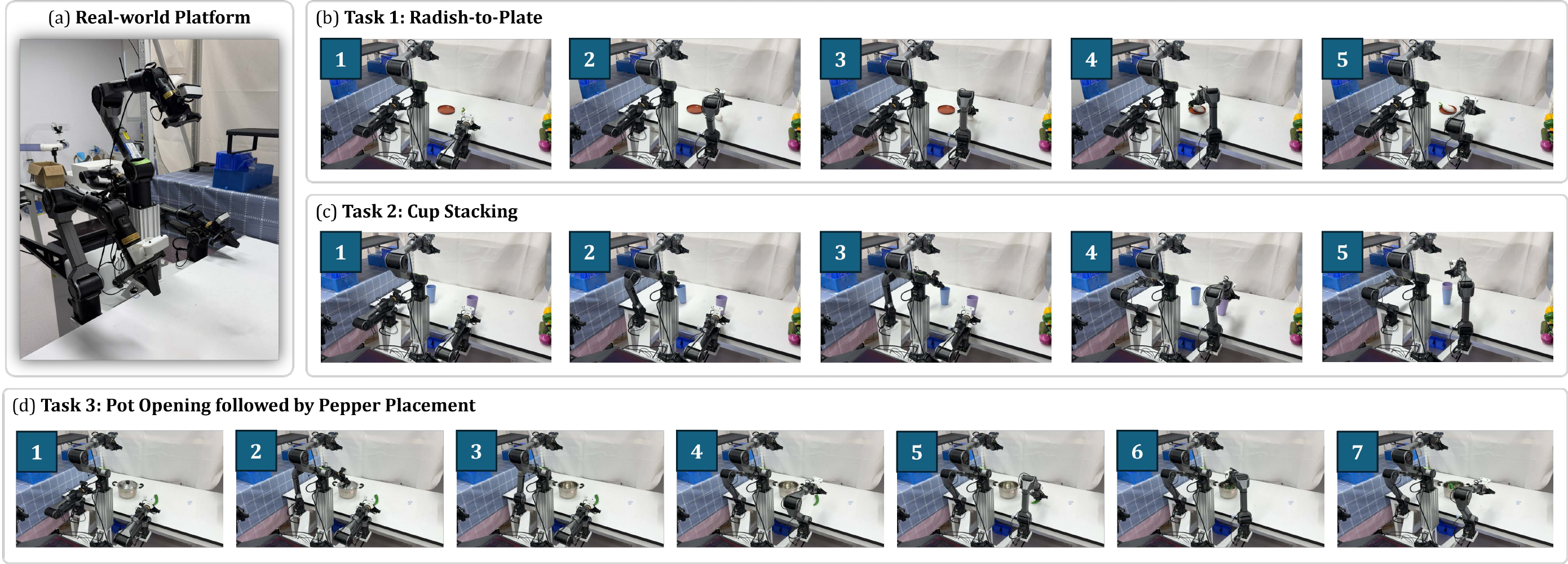}%
    }{%
        \fbox{\parbox[c][0.105\textheight][c]{\dimexpr\textwidth-2\fboxsep-2\fboxrule\relax}{%
            \centering Cobot Magic platform and recorded task sequences\\[3pt]
            \texttt{figures/3.2.pdf}}}%
    }
    \caption{
    \textbf{Real-world evaluation on Cobot Magic.}
    (a) Physical manipulation platform with the wrist-mounted scene-view camera.
    (b)--(d) Representative execution sequences for radish-to-plate,
    cup stacking, and pot opening followed by pepper placement, respectively.
    The third task consists of two stages: removing the pot lid and subsequently
    placing the pepper into the pot.
    Frames progress from left to right within each sequence.
    The shown frames are captured from an external recording viewpoint rather
    than from the camera observations supplied to the policy.
    Quantitative success rates over 50 trials per task are reported in
    Table~\ref{tab:realworld_results}.
    }
    \label{fig:realworld_experiments}
\end{figure*}

\subsection{Ablation Studies}
\label{sec:exp_ablation}

To address Q3, Table~\ref{tab:ablation_plan} isolates the geometric
mechanisms of MVG-WAM on RoboTwin~2.0.
For computationally efficient ablation, all variants are trained on the same
50 tasks using 50 clean and 200 randomized demonstrations per task and are
evaluated with shared seeds.
This reduced training regime is used only for the ablation study; the standard
RoboTwin~2.0 results in Table~\ref{tab:sim_main} use 50 clean and 500 randomized
demonstrations per task.

Hard-13 is fixed by an independent FastWAM-Joint validation pass using tasks
with SR below 90\%.
The variant without the epipolar constraint retains the same dense encoder
and retrieval budget but removes geometric restriction during cross-view
retrieval, while the variant without view routing exposes every canvas region
to the full per-view token bank.
Because the ablation series uses a reduced training regime and is evaluated
independently from the standard benchmark run, we report its corresponding
full-model result rather than substituting the score from
Table~\ref{tab:sim_main}.

The complete model improves over FastWAM-Joint by 2.61 points across all
50 tasks and 5.69 points on Hard-13.
Removing the epipolar residual, epipolar constraint, view routing, or
depth supervision consistently reduces SR.
In particular, removing metric-depth supervision decreases performance by
1.09 points on all tasks and 2.84 points on Hard-13, supporting the benefit
of metric-scale grounding for the MVG representation.
Together, these controlled comparisons support the complementary roles of
cross-view epipolar modeling, view-indexed routing, and metric-scale grounding.

\begin{table}[t]
    \centering
    \caption{
    RoboTwin~2.0 ablation success rates (\%) under the reduced
    50-clean/200-random training regime.
    Hard-13 is fixed by independent FastWAM-Joint validation.
    Rows beginning with ``w/o'' denote MVG-WAM variants with the
    corresponding component removed.
    }
    \label{tab:ablation_plan}
    \begingroup
    \footnotesize
    \setlength{\tabcolsep}{3pt}
    \renewcommand{\arraystretch}{1.06}

    \begin{tabular*}{\columnwidth}{@{\extracolsep{\fill}}lcc@{}}
        \toprule
        Variant & All 50 & Hard 13 \\
        \midrule

        \textbf{MVG-WAM (Ours)}
        & \textbf{92.05}
        & \textbf{77.65} \\

        w/o Epipolar Residual
        & 91.05
        & 75.00 \\

        w/o Epipolar Constraint
        & 90.86
        & 74.65 \\

        w/o View Routing
        & 90.93
        & 74.92 \\

        w/o Depth Supervision
        & 90.96
        & 74.81 \\

        \midrule

        FastWAM-Joint (Baseline)
        & 89.44
        & 71.96 \\

        \bottomrule
    \end{tabular*}

    \endgroup
\end{table}

\subsection{Real-World Experiments}
\label{sec:exp_realworld}

To address Q1 on a physical system, we evaluate MVG-WAM on an AgileX
Cobot Magic using radish-to-plate, cup stacking, and pot opening followed
by pepper placement.
Figure~\ref{fig:realworld_experiments} shows the platform and representative
executions.
The scene-view camera is wrist-mounted at a fixed observation angle.

We collect 100 RGB-D training and ten validation demonstrations per task
and fine-tune one three-task policy for 20k updates.
Each task is evaluated over 50 trials; interventions and safety stops count
as failures.

Table~\ref{tab:realworld_results} reports 96\%, 90\%, and 88\% SR for
MVG-WAM, yielding 137/150 successful trials (91.3\%).
Under the same evaluation protocol, $\pi_{0.5}$ achieves 92\%, 88\%, and
88\% SR, corresponding to a mean SR of 89.3\%.

The real-world results further support Q3:
removing the epipolar residual reduces mean SR by 4.0 points.

The controller executes 24-waypoint chunks at 20~Hz; a batch-size-one
policy call takes 0.81~s on an H100, including preprocessing and geometry
encoding.

\begin{table}[t]
    \centering
    \caption{
    Cobot Magic success rate (\%), 50 trials per task.
    R, C, and O denote radish-to-plate, cup stacking, and pot opening
    followed by pepper placement.
    ``w/o'' denotes without.
    }
    \label{tab:realworld_results}
    \begingroup
    \footnotesize
    \setlength{\tabcolsep}{3pt}
    \renewcommand{\arraystretch}{1.06}

    \begin{tabular*}{\columnwidth}{@{\extracolsep{\fill}}lcccc@{}}
        \toprule
        Model & R & C & O & Mean \\
        \midrule

        $\pi_{0.5}$~\cite{black2025pi05}
        & 92 & 88 & 88 & 89.3 \\

        FastWAM-Joint~\cite{yuan2026fastwam}
        & 86 & 78 & 72 & 78.7 \\

        MVG-WAM w/o Epipolar Residual
        & 94 & 86 & 82 & 87.3 \\

        \midrule

        \textbf{MVG-WAM (Ours)}
        & \textbf{96}
        & \textbf{90}
        & \textbf{88}
        & \textbf{91.3} \\

        \bottomrule
    \end{tabular*}

    \endgroup
\end{table}

\begin{figure}[t]
    \centering
    \IfFileExists{figures/4.7.pdf}{%
        \includegraphics[width=\columnwidth]{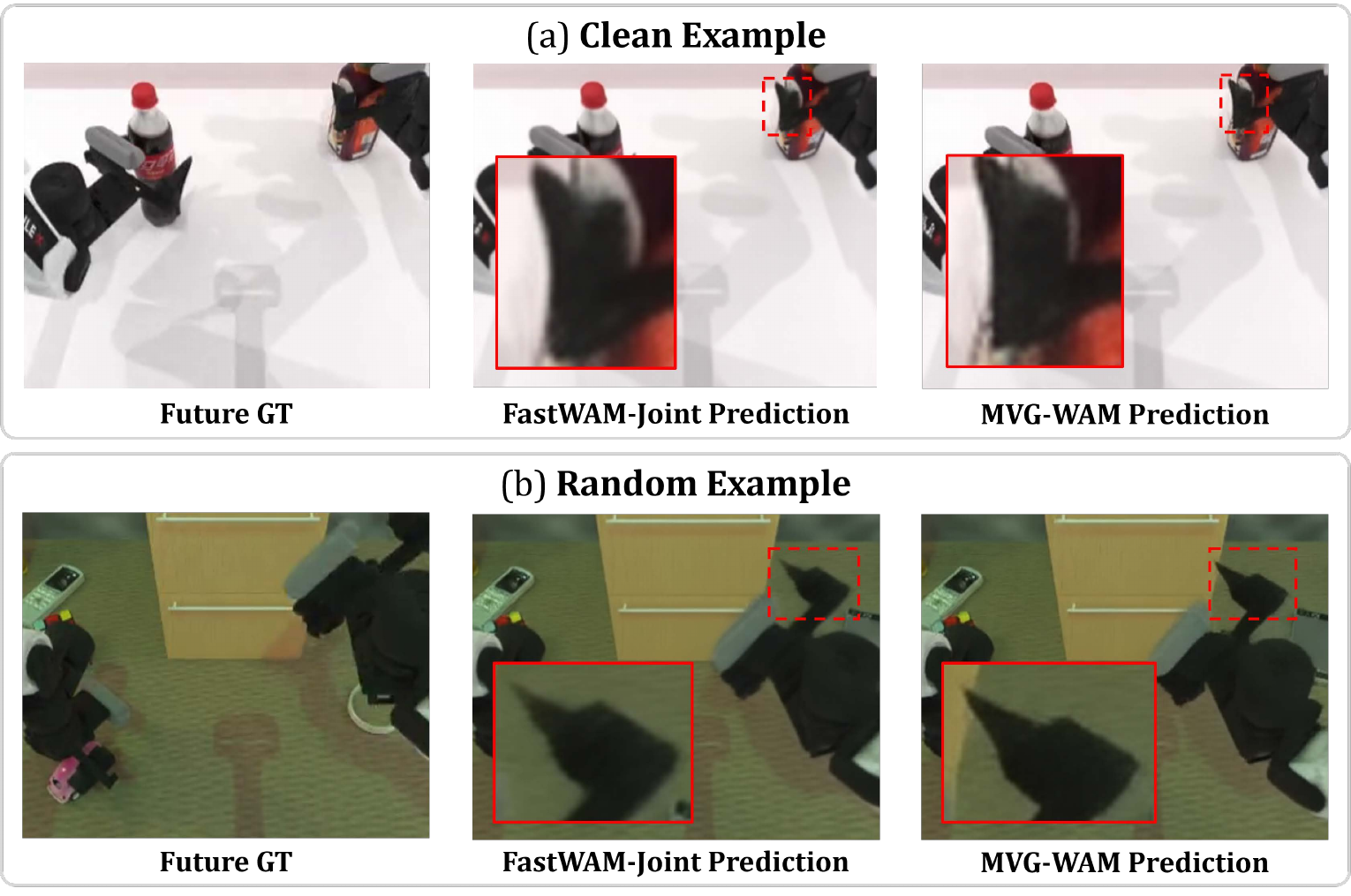}%
    }{%
        \fbox{\parbox[c][0.16\columnwidth][c]{\dimexpr\columnwidth-2\fboxsep-2\fboxrule\relax}{%
            \centering Short-horizon RGB prediction comparison\\[2pt]
            \texttt{figures/4.7.pdf}}}%
    }
    \caption{
    Qualitative short-horizon RGB prediction on RoboTwin~2.0.
    FastWAM-Joint and MVG-WAM use the same current observation and are
    compared against the same future target.
    Dashed boxes indicate gripper--object interaction regions, with enlarged
    crops shown in red.
    MVG-WAM produces more faithful local predictions, preserving sharper
    object boundaries and more coherent gripper--object geometry.
    }
    \label{fig:geometry_analysis}
\end{figure}

\begin{table}[t]
    \centering
    \caption{
    Short-horizon RGB and teacher-forced depth diagnostics on RoboTwin~2.0,
    averaged equally over Clean and Random domains.
    A dash denotes not applicable: FastWAM-Joint has no trained native depth head.
    Starred GT entries are analytical self-comparisons; CV and coverage are reprojection references.
    }
    \label{tab:geometry_metrics}
    \begingroup
    \footnotesize
    \setlength{\tabcolsep}{2.1pt}
    \renewcommand{\arraystretch}{1.06}
    \begin{tabular*}{\columnwidth}{@{\extracolsep{\fill}}lcccccc@{}}
        \toprule
        Model / ref.
        & \shortstack{MAE$\downarrow$\\(cm)}
        & \shortstack{AbsRel$\downarrow$\\(\%)}
        & \shortstack{CV$\downarrow$\\(cm)}
        & \shortstack{Cov.\\(\%)}
        & \shortstack{PSNR$\uparrow$\\(dB)}
        & SSIM$\uparrow$ \\
        \midrule

        FastWAM-Joint
        & \multicolumn{1}{c}{---}
        & \multicolumn{1}{c}{---}
        & \multicolumn{1}{c}{---}
        & \multicolumn{1}{c}{---}
        & 28.05
        & 0.842 \\

        \textbf{MVG-WAM (Ours)}
        & 2.12
        & 5.80
        & 1.68
        & 16.70
        & 31.79
        & 0.914 \\

        GT reference
        & $0.00^{*}$
        & $0.00^{*}$
        & 0.25
        & 19.00
        & $\infty^{*}$
        & $1.000^{*}$ \\

        \bottomrule
    \end{tabular*}
    \endgroup
\end{table}

\subsection{Multi-View Geometric Analysis}
\label{sec:exp_analysis}

To address Q4, we evaluate short-horizon RGB prediction and
teacher-forced (TF) metric-depth readout on 625 clips from 125 trajectories
in the existing RoboTwin~2.0 validation split.

RGB prediction uses only the current observation, proprioception, and
instruction context.
The depth readout is evaluated separately from noisy future latents and
therefore serves as a diagnostic of the learned representation rather than
current-observation-only depth forecasting.
For compactness, Table~\ref{tab:geometry_metrics} reports the equal average
of the previously computed Clean and Random task-macro scores.
FastWAM-Joint has no trained native depth head, so its corresponding depth
metrics are not applicable.

MVG-WAM improves average RGB PSNR by 3.74~dB and SSIM by 0.072 over
FastWAM-Joint.
Figure~\ref{fig:geometry_analysis} provides a qualitative comparison of the
corresponding short-horizon predictions.
The improvement is particularly visible around gripper--object interaction
regions, where MVG-WAM better preserves local object boundaries and
gripper--object structure.
The enlarged crops highlight these manipulation-critical regions while both
methods are compared against the same future target.

For the TF depth diagnostic, we corrupt future target latents at a fixed
video diffusion timestep $\tau_v=400$ before reading out metric depth.
Under this setting, the averaged TF depth readout of MVG-WAM reaches
2.12~cm MAE, 5.80\% AbsRel, and 1.68~cm cross-view $z$ error.
Its reprojection coverage is 16.70\%, compared with 19.00\% for the GT
reference; coverage is therefore interpreted together with reprojection
error rather than as a stand-alone accuracy measure.
These depth diagnostics indicate that the denoising representation retains
metric geometric information across views and future horizons.
Since FastWAM-Joint has no trained depth head, they are not interpreted as
a direct depth-prediction improvement over the baseline.

\section{Conclusion}
\label{sec:conclusion}

We presented MVG-WAM, a video-based world--action model that treats
synchronized camera observations as geometrically related projections
of one physical world.
By combining an epipolar-constrained global state with camera-aware routing,
the model integrates cross-view evidence while retaining view-indexed
geometric context.
Future metric-depth supervision metrically grounds this geometry-aware
representation as the scene evolves, without requiring depth decoding
during action rollout.

Without additional embodied pretraining, MVG-WAM achieves average success
rates of 99.1\% on LIBERO and 92.07\% on RoboTwin~2.0.
Real-world evaluation further achieves 91.3\% success across 150 trials
on Cobot Magic.
Together, these results demonstrate that explicitly organizing multi-view
geometry provides an effective representation for world--action modeling
in robotic manipulation.


\balance
\bibliographystyle{IEEEtran}
\bibliography{references}
\end{document}